\documentclass{bioinfo}
\copyrightyear{2016} \pubyear{2016}

\access{}
\appnotes{}
\usepackage{epstopdf}
\usepackage[ruled,vlined]{algorithm2e}
\usepackage{algpseudocode}
\usepackage{multirow}

\begin{document}
\firstpage{1}

\subtitle{}

\title[\textsc{ProtNN}: Fast and Accurate Nearest Neighbor Protein Function Prediction]{\textsc{ProtNN}: Fast and Accurate Nearest Neighbor Protein Function Prediction based on Graph Embedding in Structural and Topological Space}
\author[Dhifli \textit{et~al}]{Wajdi Dhifli$^{1}$, and Abdoulaye Banir\'e Diallo\,$^{1,\ast}$}
\address{$^{1}$Department of Computer Science, University of Quebec At Montreal, PO box 8888, Downtown station, Montreal, Qc, Canada, H3C 3P8.}

\corresp{$^\ast$To whom correspondence should be addressed.}

\history{}

\editor{}

\abstract{\textbf{Motivation:} Studying the function of proteins is important for understanding the molecular mechanisms of life. The number of publicly available protein structures has increasingly become extremely large. Still, the determination of the function of a protein structure remains a difficult, costly, and time consuming task. The difficulties are often due to the essential role of spatial and topological structures in the determination of protein functions in living cells.\\
\textbf{Results:} In this paper, we propose \textsc{ProtNN}, a novel approach for protein function prediction. Given an unannotated protein structure and a set of annotated proteins, \textsc{ProtNN} finds the nearest neighbor annotated structures based on protein-graph pairwise similarities. Given a query protein, \textsc{ProtNN} finds the nearest neighbor reference proteins based on a graph representation model and a pairwise similarity between vector embedding of both query and reference protein-graphs in structural and topological spaces. 
\textsc{ProtNN} assigns to the query protein the function with the highest number of votes across the set of $k$ nearest neighbor reference proteins, where $k$ is a user-defined parameter. Experimental evaluation demonstrates that \textsc{ProtNN} is able to accurately classify several datasets in an extremely fast runtime compared to state-of-the-art approaches. We further show that \textsc{ProtNN} is able to scale up to a whole PDB dataset in a single-process mode with no parallelization, with a gain of thousands order of magnitude of runtime compared to state-of-the-art approaches.\\
\textbf{Availability:} An implementation of \textsc{ProtNN} as well as the experimental datasets are available at \href{https://sites.google.com/site/wajdidhifli/softwares/protnn}{https://sites.google.com/site/wajdidhifli/softwares/protnn}.\\
\textbf{Contact:} \href{diallo.abdoulaye@uqam.ca}{diallo.abdoulaye@uqam.ca}
}

\maketitle

\section{Introduction}
Proteins are ubiquitous in the living cells. They play key roles in the functional and evolutionary machinery of species. Studying protein functions is paramount for understanding the molecular mechanisms of life. 
High-throughput technologies are yielding millions of protein-encoding sequences that currently lack any functional characterization \citep{Brenner2000, Lee2007, Molloy2014} 
The number of proteins in the Protein Data Bank (PDB) \citep{Berman_2000} has more than tripled over the last decade. Alternative databases such as SCOP \citep{Andreeva_2008} and CATH \citep{Sillitoe_2015} are undergoing the same trend. However, the determination of the function of protein structures remains a difficult, costly, and time consuming task. Manual protein functional classification methods are no longer able to follow the rapid increase of data. Accurate computational and machine learning tools present an efficient alternative that could offer considerable boosting to meet the increasing load of data.

Proteins are composed of complex three-dimensional folding of long chains of amino acids. This spatial structure is an essential component in protein functionality and is thus subject to evolutionary pressures to optimize the inter-residue contacts that support it \citep{Meysman_2015}. Existing computational methods for protein function prediction try to simulate biological phenomena that define the function of a protein. The most conventional technique is to perform a similarity search between an unknown protein and a reference database of annotated proteins with known functions. The query protein is assigned with the same functional class of the most similar (based on the sequence or the structure) reference protein. There exists several classification methods based on the protein sequence ($e.g.$ Blast \citep{Altschul_1990}, ...); or on the protein structure ($e.g.$ Combinatorial Extension \citep{Bourne_1998}, Sheba \citep{Lee_2000}, FatCat \citep{Ye_2003}, Fragbag \citep{Budowski-Tal23022010}, ...). These methods rely on the assumption that proteins sharing the most common sites are more likely to share functions. This classification strategy is based on the hypothesis that structurally similar proteins could share a common ancestor \citep{Borgwardt_2005}. Another popular approach for protein functional classification is to look for relevant substructures (also so-called motifs) among proteins with known functions, then use them as features to identify the function of unknown proteins. Such motifs could be discriminative \citep{Zhu_2012}, representative \citep{Dhifli_2014}, cohesive \citep{Meysman_2015}, $etc$. Each of the mentioned protein functional classification approaches suffers different drawbacks. Sequence-based classification do not incorporate spatial information of amino acids that are not contiguous in the primary structure but interconnected in 3D space. This makes them less efficient in predicting the function for structurally similar proteins with low sequence similarity (remote homologues). Both structure and substructure-based classification techniques do incorporate spatial information in function prediction which makes them more efficient than sequence-based classification. However, such consideration makes these methods subject to the \textit{"no free lunch"} principle \citep{Dolpert_1997}, where the gain in accuracy comes with an offset of computational cost. Hence, it is essential to find an efficient way to incorporate 3D-structure information with low computational complexity.

In this paper, we present \textsc{ProtNN}, a novel approach for function prediction of protein 3D-structures. \textsc{ProtNN} incorporates protein 3D-structure information via the combination of a rich set of structural and topological descriptors that guarantee an informative multi-view representation of the structure that considers spatial information through different dimensions. Such a representation transforms the complex protein 3D-structure into an attribute-vector of fixed size allowing computational efficiency. For classification, \textsc{ProtNN} assigns to a query protein the function with the highest number of votes across the set of its $k$ most similar reference proteins, where $k$ is a user-defined parameter. Experimental evaluation shows that \textsc{ProtNN} is able to accurately classify different benchmark datasets with a gain of up to 47x of computational cost compared to gold standard approaches from the literature such as  Combinatorial Extension \citep{Bourne_1998}, Sheba \citep{Lee_2000}, FatCat \citep{Ye_2003} and others. We further show that \textsc{ProtNN} is able to scale up to a PDB-wide dataset in a single-process mode with no parallelization, where it outperformed state-of-the-art approaches with thousands order of magnitude in runtime on classifying a 3D-structure against the entire PDB.

\begin{methods}
\section{Methods}
\subsection{Graph Representation of Protein 3D-Structures}
A crucial step in computational studies of protein 3D-structures is to look for a convenient representation of their spatial conformations. Graphs represent the most appropriate data structures to model the complex structures of proteins. In this context, a protein 3D-structure can be seen as a set of elements (amino acids and atoms) that are interconnected through chemical interactions \citep{Borgwardt_2005,Dhifli_2014,Meysman_2015}. These interactions are mainly:
\begin{itemize}
\item[-] Covalent bonds between atoms sharing pairs of valence electrons,
\item[-] Ionic bonds of electrostatic attractions between oppositely charged components,
\item[-] Hydrogen bonds between two partially negatively charged atoms sharing a partially positively charged hydrogen,
\item[-] Hydrophobic interactions where hydrophobic amino acids in the protein closely associate their side chains together,
\item[-] Van der Waals forces which represent transient and weak electrical attraction of one atom for another when electrons are fluctuating.
\end{itemize}
These chemical interactions are supposed to be the analogues of graph edges. Figure \ref{fig:protein_to_graph} shows a real example of the human hemoglobin protein and its graph representation. The Figure shows clearly that the graph representation preserves the overall structure of the protein and its components. 
\begin{figure*}
\centerline{\includegraphics[width=0.7\textwidth]{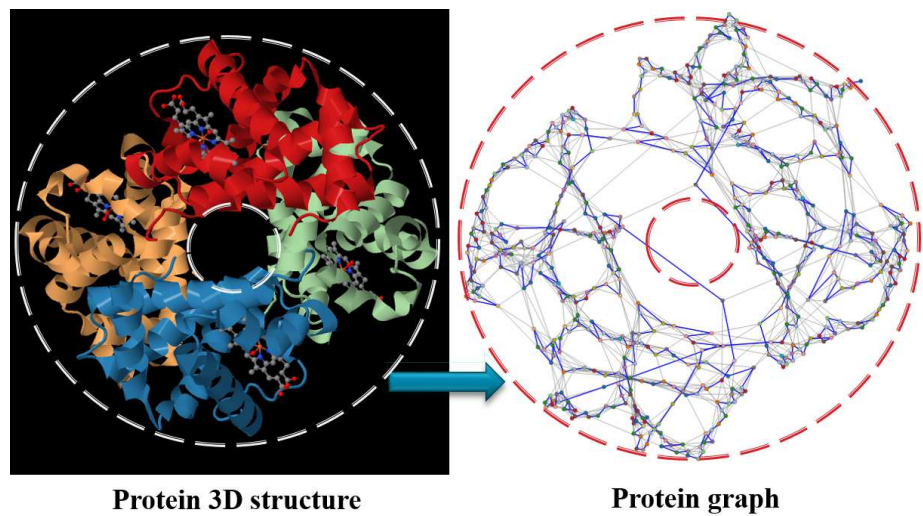}}
\caption{The human hemoglobin protein 3D-structure (PDBID: 1GZX) and its corresponding graph representation. Nodes and edges represent, respectively, amino acids from the structure and links between them. Blue edges represent links from the primary structure and gray edges are spatial links between distant amino acids.}\label{fig:protein_to_graph}
\end{figure*}
\subsubsection{Protein Graph Model}
Let $G$ be a graph consisting of a set of nodes $V$ and edges $E$. $L$ is a label function that associates a label $l$ to each node in $V$. Each node of $G$ represents an amino acid from the 3D-structure, and is labeled with its corresponding amino acid type. Let $\Delta$ be a function that computes the euclidean distance between pairs of nodes $\Delta(u,v), \forall u, v \in V$, and $\delta$ a distance threshold. Each node in $V$ is defined by its 3D coordinates in ${\rm I\!R}^3$, and both $\Delta$ and $\delta$ are expressed in angstroms ($\AA$). Two nodes $u$ and $v$ ($\forall u, v \in V$) are linked by an edge $e(u, v)\in E$, if the distance between their $C_\alpha$ atoms is below or equal to $\delta$. Formally, the adjacency matrix $A$ of $G$ is defined as follows:
\begin{equation}\label{eq:protein_to_graph}
A_{u,v}= \begin{cases}
	 1, \emph{  }if\emph{  } \Delta(C_{\alpha_u}, C_{\alpha_v})\leq \delta \\
	 0, \emph{  }otherwise 
\end{cases}
\end{equation}
%

\subsection{Structural and Topological Embedding of Protein Graphs}
\subsubsection{Graph Embedding}
Graph-based representations are broadly used in multiple application fields including bioinformatics \citep{Borgwardt_2005,Gibert_2010,Dhifli_2014}. However, they suffer major drawbacks with regards to processing tools and runtime. Graph embedding into vector spaces is a very popular technique to overcome both drawbacks \citep{Gibert_2010}. It aims at providing a feature vector representation for every graph, allowing to bridge the gap between the representational power of graphs, the rich set of algorithms that are available for feature-vector representations, and the need for rapid processing algorithms to handle the massively available biological data. 
In \textsc{ProtNN}, each protein 3D-structure is represented by a graph according to Equation \ref{eq:protein_to_graph}. Then, each graph is embedded into a vector of structural and topological features under the assumption that structurally similar graphs should give similar structural and topological feature-vectors. In such manner, \textsc{ProtNN} guarantees accuracy and computational efficiency. It is worth noting that even though structurally similar graphs should have similar topological properties, \textsc{ProtNN} similarity should not necessarily give the same results of structure matching (as in structural alignment). But, it should enrich it since \textsc{ProtNN} considers even hidden similarities (like graph density and energy) that are not considered in structure matching. 
%
%
%
\subsubsection{Structural and Topological Attributes}\label{subsec:attributes}
In \textsc{ProtNN}, the pairwise similarity between two protein-graphs is measured by the distance between their vector representations. In order to avoid the loss of structural information in the embedding, and to guarantee \textsc{ProtNN} accuracy, we use a set of structural and topological attributes from the literature that have shown to be interesting and efficient in describing connected graphs \citep{Leskovec_2005,Li_2012}. It is important to mention that this list could be extended as needed. In the following is the list of structural and topological attributes used in \textsc{ProtNN}:
\begin{enumerate}
\item [A1-] \textbf{Number of nodes}: The total number of nodes of the graph, $|V|$.
\item [A2-] \textbf{Number of edges}: The total number of edges of the graph, $|E|$.
\item [A3-] \textbf{Average degree}: The degree of a node $u$, denoted $deg(u)$, is the number of its adjacent nodes. The average degree of a graph $G$ is the average of all $deg(u)$, $\forall u\in G$. Formally: $ deg(G) = \frac{1}{\mid V\mid} \sum^{\mid V\mid}_{i=1} deg(u_i)$. 
\item [A4-] \textbf{Density}: The density of a graph $G=(V, E)$ measures how many edges are in $E$ compared to the number of maximum possible edges between the nodes in $V$. Formally: $ den(G) = \frac{2 \mid E\mid}{(\mid V\mid\ast (\mid V\mid -1))}$.
\item [A5-] \textbf{Average clustering coefficient}: The clustering coefficient of a node $u$, denoted $c(u)$, measures how complete the neighborhood of $u$ is, $c(u)= \frac{2 e_u}{k_u (k_u - 1)}$ where $k_u$ is the number of neighbors of $u$ and $e_u$ is the number of connected pairs of neighbors. 
The average clustering coefficient of a graph $G$, is given as the average value over all of its nodes. Formally: $C(G)= \frac{1}{\mid V\mid} \sum_{i=1}^{\mid V\mid} c(u_i)$.
\item [A6-] \textbf{Average effective eccentricity}: For a node $u$, the effective eccentricity represents the maximum length of the shortest paths between $u$ and every other node $v$ in $G$, $e(u) = max\lbrace d(u,v) : v\in V, u\neq v\rbrace $, where $d(u,v)$ is the length of the
shortest path from $u$ to $v$. 
The average effective eccentricity is defined as $Ae(G)= \frac{1}{\mid V\mid}\sum_{i=1}^{\mid V\mid} e(u_i)$.
\item [A7-] \textbf{Effective diameter}: It represents the maximum value of effective eccentricity over all nodes in the graph $G$, $i.e.$, $diam(G) = max\lbrace e(u)\mid u\in V\rbrace$ where $e(u)$ represents the effective eccentricity of $u$ as defined above.
\item [A8-] \textbf{Effective radius}: It represents the minimum value of effective eccentricity over all nodes of $G$, $rad(G) = min\lbrace e(u)\mid u\in V\rbrace$.
\item [A9-] \textbf{Closeness centrality}: The closeness centrality measures how fast information spreads from a given node to other reachable nodes in the graph. For a node $u$, it represents the reciprocal of the average shortest path length between $u$ and every other reachable node in the graph $G$, $C_c(u) = \frac{{\mid V\mid}-1}{\sum_{v\in \lbrace V\setminus u\rbrace} d(u,v)}$ where $d(u,v)$ is the length of the shortest path between the nodes $u$ and $v$. For $G$, we consider the average value of closeness centrality of all its nodes, $C_c(G) = \frac{1}{\mid V\mid} \sum_{i=1}^{\mid V\mid} C_c(u_i)$.
\item [A10-] \textbf{Percentage of central nodes}: It is the ratio of the number of central nodes from the number of nodes in the graph. A node $u$ is central if the value of its eccentricity is equal to the effective radius of the graph, $e(u) = rad(G)$.
\item [A11-] \textbf{Percentage of end points}: It represents the ratio of the number of nodes with $deg(u) = 1$ from the total number of nodes of $G$. 
\item [A12-] \textbf{Number of distinct eigenvalues}: The adjacency matrix $A$ of $G$ has a set of eigenvalues. We count the number of distinct eigenvalues of $A$.
\item [A13-] \textbf{Spectral radius}: Let $\leftthreetimes_1 , \leftthreetimes_2 , ..., \leftthreetimes_m$ be the set of eigenvalues of the adjacency matrix $A$ of $G$. The spectral radius of $G$, denoted $\rho(G)$, represents the largest magnitude eigenvalue, $i.e.$, $\rho(G) = max(\mid\leftthreetimes_i\mid)$ where $i\in \lbrace 1, .., m\rbrace$.
\item [A14-] \textbf{Second largest eigenvalue}: The value of the second largest eigenvalue.
\item [A15-] \textbf{Energy}: The energy of an adjacency matrix $A$ of a graph $G$ is defined as the squared sum of the eigenvalues of $A$. Formally: $E(G) = \sum^m_{i=1}\leftthreetimes_i^2$.
\item [A16-] \textbf{Neighborhood impurity}: For a node $u$ having a label $L(u)$ and a neighborhood $N(u)$, it is defined as $ImpDeg(u) = \mid L(v): v \in N(u), L(u)\neq L(v)\mid$. The neighborhood impurity of $G$ is the average $ImpDeg$ over all nodes.
\item [A17-] \textbf{Link impurity}: An edge $\{u,v\}$ is considered to be impure if $L(u)\neq L(v)$. The link impurity of a graph $G$ with $|E|$ edges is defined as: $\frac{\mid\{u,v\}\in E: L(u)\neq L(v)\mid}{\mid E\mid}$.
\item [A18-] \textbf{Label entropy}: It measures the uncertainty of labels. For a graph $G$ of $k$ labels, it is defined as $E(G)= -\sum_{i=1}^k p(l_i)\textit{ log }p(l_i)$, where $l_i$ is the $i^{th }$ label.
\end{enumerate}

\subsection{\textsc{ProtNN}: Nearest Neighbor Protein Functional Classification}
The general classification pipeline of \textsc{ProtNN} can be described as follows: first a preprocessing is performed on the reference protein database $\Omega$ in which a graph model $G_P$ is created for each reference protein $P$, $\forall P\in\Omega$, according to Equation \ref{eq:protein_to_graph}. A structural and topological description vector $V_P$ is created for each graph model $G_P$, by computing the corresponding values of each of the structural and topological attributes described in Section \ref{subsec:attributes}. The resulting matrix $M_\Omega = \bigcup V_P$, $\forall P\in\Omega$, represents the preprocessed reference database that is used for prediction in \textsc{ProtNN}. In order to guarantee an equal participation of all used attributes in the classification, a min-max normalization ($x_{normalized} = \frac{x - min}{max-min}$, where $x$ is an attribute value, $min$ and $max$ are the minimum and maximum values for the attribute vector) is applied on each attribute of $M_\Omega$ independently such that no attribute will dominate in the prediction. It is also worth mentioning that for real world applications $M_\Omega$ is only computed once, and can be incrementally updated with other attributes as well as newly added protein 3D-structures with no need to recompute the attributes for the entire set. This guarantees a high flexibility and easy extension of \textsc{ProtNN} in real world application. The prediction step in \textsc{ProtNN} is described in Algorithm \ref{alg:main_algo}. 
In prediction, a query protein 3D-structure $Q$ with an unknown function, is first transformed into its corresponding graph model $G_Q$. The structural and topological attributes are computed for $G_Q$ forming its query description vector $V_Q$. The query protein $Q$ is scanned against the entire reference database $\Omega$, where the distance between $V_Q$ and each of the reference vectors $\forall V_P \in M_\Omega$ is computed and stored in $Vdist_Q$, with respect to a distance measure. The $k$ most similar reference proteins \textit{NN}$_Q^k$ are selected, and the query protein $Q$ is predicted to exert the function with the highest number of votes across the set of \textit{NN}$_Q^k$ reference proteins, where $k$ is a user-defined number of nearest neighbors.
\begin{algorithm}[!htpd]
\label{alg:main_algo}
\caption{\textsc{ProtNN} (The prediction step)}
\KwData{$Q$: Query protein 3D-structure, $M_\Omega$: Description matrix of the reference database of protein 3D-structures, $k$: number of similar}
\KwResult{$C_Q$: Functional class of $Q$}
\Begin{
$G_Q \leftarrow$ create a graph model for $Q$ according to Equation \ref{eq:protein_to_graph}\;
$V_Q \leftarrow$ $G_Q$ is embedded into a vector $V$ using the attributes\;
\textit{NN}$_Q^k \leftarrow \emptyset $ \;
\textit{Vdist}$_Q \leftarrow \emptyset $ \;
\ForEach{($V_P$ in $M_\Omega$)}{
	\textit{Vdist}$_Q$[$P$]$ \leftarrow$ distance($V_Q$, $V_P$); \Comment{The distance between vectors of query protein $Q$ and the reference protein $P$.}
}
\textit{NN}$_Q^k \leftarrow$ \textit{Top}$_k$(\textit{Vdist}$_Q$); \Comment{Select the $k$ nearest reference protein neighbors}\\
$C_Q \leftarrow$ The functional class with the highest number of votes across the set of \textit{NN}$_Q^k$ reference proteins\;
}
\end{algorithm}
\end{methods}

\section{Experiments}
\subsection{Datasets}
\subsubsection{Benchmark Datasets}
To assess the classification performance of \textsc{ProtNN}, we performed an experiment on six well-known benchmark datasets of protein structures that have previously been used in \citep{Jin_2009,Jin_2010,Fei_2010,Zhu_2012}. Each dataset is composed of positive protein examples that are from a selected protein family, and negative protein examples that are randomly sampled from the PDB \citep{Berman_2000}. Table \ref{tab:datasets} summarizes the characteristics of the six datasets. SCOP ID, Family name, Pos., and Neg. correspond respectively to the identifier of the protein family in SCOP \citep{Andreeva_2008}, its name, and the number of positive and negative examples. The selected positive protein families are Vertebrate phospholipase A2, G-protein family, C1-set domains, C-type lectin domains, and protein kinases, catalytic subunits.

\textbf{Vertebrate phospholipase A2:} Phospholipase A2 are enzymes from the class of hydrolase, which release the fatty acid from the hydroxyl of the carbon 2 of glycerol to give a phosphoglyceride lysophospholipid. They are located in most mammalian tissues.

\textbf{G-proteins:} G-proteins are also known as guanine nucleotide-binding proteins. These proteins are mainly involved in transmitting chemical signals originating from outside a cell into the inside of it. G-proteins are able to activate a cascade of further signaling events resulting a change in cell functions. They regulate metabolic enzymes, ion channels, transporter, and other parts of the cell machinery, controlling transcription, motility, contractility, and secretion, which in turn regulate diverse systemic functions such as embryonic development, learning and memory, and homeostasis.

\textbf{C1-set domains:} The C1-set domains are immunoglobulin-like domains, similar in structure and sequence. They resemble the antibody constant domains. They are mostly found in molecules involved in the immune system, in the major histocompatibility complex class I and II complex molecules, and in various T-cell receptors.

\textbf{C-type lectin domains:} Lectins occur in plants, animals, bacteria and viruses. The C-type (Calcium-dependent) lectins are a family of lectins which share structural homology in their high-affinity carbohydrate-recognition domains. This dataset involves groups of proteins playing divers functions including cell-cell adhesion, immune response to pathogens and apoptosis.

\textbf{Proteasome subunits:} Proteasomes are critical protein complexes that primarily function to breakdown unneeded or damaged proteins. They are located in the nucleus and cytoplasm. The proteasome recycles damaged and misfolded proteins as well as degrades short-lived regulatory proteins. As such, it is a critical regulator of many cellular processes, including the cell cycle, DNA repair, signal transduction, and the immune response.

\textbf{Protein kinases, catalyc subunits:} Protein kinases, catalytic subunit play a role in various cellular processes, including division, proliferation, apoptosis, and differentiation. They are mainly proteins that modify other ones by chemically adding phosphate groups to them. This usually results in a functional change of the target protein by changing enzyme activity, cellular location, or association with other proteins. The catalytic subunits of protein kinases are highly conserved, and several structures have been solved, leading to large screens to develop kinase-specific inhibitors for the treatments of a number of diseases.

\begin{table}[!t]
\processtable{Characteristics of the experimental datasets. SCOP ID: identifier of protein family in SCOP, Pos.: number of positive examples, Neg.: number of negative examples.\label{tab:datasets}}
{\begin{tabular}{lllll}
\hline\noalign{\smallskip}
\textbf{Dataset} & \textbf{SCOP ID} & \textbf{Family name} & \textbf{Pos.} & \textbf{Neg.}\\ 
\noalign{\smallskip}
\hline
\noalign{\smallskip}
DS1	& 48623	& Vertebrate phospholipase A2	& 29 & 	29\\ 
DS2	& 52592 & G-proteins & 33	& 33\\ 
DS3	& 48942	& C1-set domains & 38	& 38\\ 
DS4	& 56437 & C-type lectin domains & 38	& 38\\ 
DS5	& 56251	& Proteasome subunits & 35	& 35\\ 
DS6	& 88854	& Protein kinases, catalyc subunits	& 41	& 41\\\hline
\end{tabular}}{}
\end{table}

\subsubsection{The Protein Data Bank}\label{subsec:PDB}
In order to assess the scalability of \textsc{ProtNN} to large scale real-world applications, we evaluate the runtime of our approach on the entire Protein Data Bank (PDB) \citep{Berman_2000} which contains the updated list of all known protein 3D-structures. We use 94126 structure representing all the available protein 3D-structures in the PDB by the end of July 2014.

\begin{figure*}[!htpb]
\centerline{\includegraphics[width=0.95\textwidth]{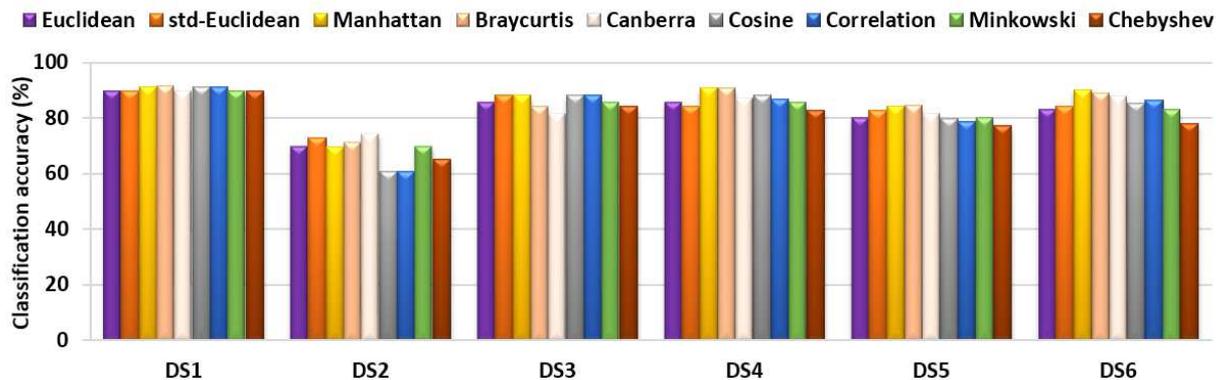}}
\caption{Classification accuracy of \textsc{ProtNN} using different distance measures ($k$=1).}\label{fig:accuracy_distance_measures}
\end{figure*}

\subsection{Protocol and Settings}
Experiments were conducted on a \textit{CentOS} Linux workstation with an Intel core-i7 CPU at 3.40 GHz, and 16.00 GB of RAM. All the experiments are performed in a single process mode with no parallelization. To transform protein into graph, we used a $\delta$ value of 7$\AA$. The evaluation measure is the classification accuracy, and the evaluation technique is Leave-One-Out (LOO) where each dataset is used to create $N$ classification scenarios, where $N$ is the number of proteins in the dataset. In each scenario, a reference protein is used as a query instance and the rest of the dataset is used as reference. The aim is to correctly predict the class of the query protein. The classification accuracy for each dataset is averaged over results of all the $N$ evaluations.

\section{Results and Discussion}
\subsection{\textsc{ProtNN} Classification Results}
\subsubsection{Results Using Different Distance Measures}
The classification algorithm of \textsc{ProtNN} supports any user-defined distance measure. In this section, we study the effect of varying the distance measure on the classification accuracy of \textsc{ProtNN}. We fixed $k$=1, and we used nine different well-known distance measures namely \textit{Euclidean}, \textit{standardized Euclidean} (std-euclidean), \textit{Cosine}, \textit{Manhattan}, \textit{Correlation}, \textit{Minkovski}, \textit{Chebyshev}, \textit{Canberra}, and \textit{Braycurtis}. See \citep{Sergio_2015} for a formal definition of these measures. Figure \ref{fig:accuracy_distance_measures} shows the obtained results. 

Overall, varying the distance measure did not significantly affect the classification accuracy of \textsc{ProtNN} on the six datasets. 
Indeed, the standard deviation of the classification accuracy of \textsc{ProtNN} with each distance measure did not exceed 3\% on the six datasets. 
A ranking based on the average classification accuracy over the six datasets suggests the following descending order: (1) Manhattan, (2) Braycurtis, (3) std-Euclidean, (4) Canberra, (5) Cosine, (6) Euclidean - Minkowski, (8) Correlation, (9) Chebyshev.

\subsubsection{Results Using Different Numbers of Nearest Neighbors}
\begin{figure}[!htpb]
\centerline{\includegraphics[width=0.5\textwidth]{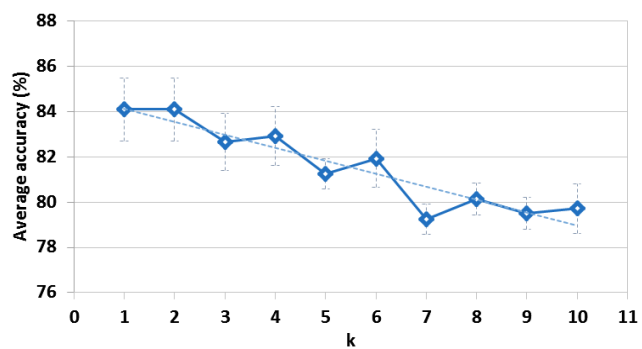}}
\caption{Tendancy of the average accuracy of \textsc{ProtNN} for each value of $k\in$ [1,10] over the six datasets and using each of the top-five distance measures. The dashed line represents the linear tendancy of the results.}\label{fig:best_k_average}
\end{figure}

In the following, we evaluate the classification accuracy of \textsc{ProtNN} on each of the six benchmark datasets using different numbers of nearest neighbors $k\in$ [1,10]. The same experiment is performed using each of the top-five distance measures. 
For simplicity, we only plot the average value of classification accuracy for each value of $k\in$ [1,10] over the six datasets using each of the top-five measures. Note that the standard deviation of the classification accuracy with each value of $k$ did not exceed 2\%. Figure \ref{fig:best_k_average} shows the obtained results. 
%
%
The number of nearest neighbors $k$ has a clear effect on the accuracy of \textsc{ProtNN}. 
The results suggest that the "optimal" value of $k\in$ \{1,2\}. 
The overall accuracy tendency shows that it decreases with higher values of $k$. This is due to the structural similarity that a query protein may share with other evolutionary close proteins exerting different functions. High values of $k$ engender considering too many neighbors which may causes a misclassification.
%

\subsubsection{Analysis of the Used Attributes}
In the following, we study the importance of the used attributes in order to identify the most informative ones. We follow the Recursive Feature Elimination (RFE) \citep{Guyon_2002} using \textsc{ProtNN} as the classifier. In RFE, one feature is removed at each iteration, where the remaining features are the ones that best enhance the classification accuracy. The pruning stops when no further enhancement is observed or no more features are left. The remaining features constitute the optimal subset for that context. 
\begin{table*}[!t]
\processtable{Empirical ranking of the structural and topological attributes.\label{tab:attribute_ranking}}
{\begin{tabular}{lcccccccccccccccccc}
\hline\noalign{\smallskip}
\multicolumn{1}{c}{\multirow{2}{*}{\textbf{Data}}} & \multicolumn{18}{c}{\textbf{Attributes}} \\\cline{2-19}\noalign{\smallskip}
\multicolumn{1}{c}{} & \textbf{A1} & \textbf{A2} & \textbf{A3} & \textbf{A4} & \textbf{A5} & \textbf{A6} & \textbf{A7} & \textbf{A8} & \textbf{A9} & \textbf{A10} & \textbf{A11} & \textbf{A12} & \textbf{A13} & \textbf{A14} & \textbf{A15} & \textbf{A16} & \textbf{A17} & \textbf{A18} \\\hline\noalign{\smallskip}
DS1 	& 0 & 2	& 1 & 6 & 2 & 1 & 5 & 9 & 10 & 1 & 2 & 8 & 6 & 13 & 16& 17 & 12 & 17 \\
DS2 	& 8 & 12 & 15 & 16 & 18 & 4 & 9& 16 & 23 & 17 & 9 & 21 & 11 & 14 & 25 & 17 & 23 & 9 \\
DS3 	& 8 & 13 & 2 & 6 & 17 & 10 & 16 & 11 & 11 & 4 & 8 & 18 & 21 & 2 & 21 & 23 & 9 & 18 \\
DS4 	& 4 & 7 & 21 & 17 & 20 & 6 & 11 & 17 & 16 & 7 & 2 & 14 & 21 & 22 & 20 & 21 & 24 & 17 \\
DS5 	& 12 & 12 & 8 & 10 & 12 & 5 & 7& 7& 17& 17 & 7& 23 & 23 & 9& 20 & 9  & 19 & 18 \\
DS6 	& 5 & 11 & 9  & 8 & 11 & 6  & 14 & 14 & 13  & 6  & 1  & 17 & 14   & 18 & 24 & 10 & 17 & 13 \\\hline\noalign{\smallskip}
\textbf{Total}	& 37 & 57 & 56 & 63 & 80 & 32 & 62 & 74 & 90  & 52 & 29 & \textbf{101}& \textbf{96} & 78 & \textbf{126}  & \textbf{97} & \textbf{104}& 92 \\\hline\noalign{\smallskip}
\textbf{Score}	& 0.25 & 0.38 & 0.37 & 0.42 & 0.53 & 0.21 & 0.41 & 0.49 & 0.6 & 0.35 & 0.19 & \textbf{0.67} & \textbf{0.64} & 0.52 & \textbf{0.84} & \textbf{0.65} & \textbf{0.69} & 0.61 \\\hline\noalign{\smallskip}
\textbf{Rank}	& 16 & 13 & 14 & 11 & 8& 17 & 12 & 10 & 7 & 15 & 18 & \textbf{3}  & \textbf{5}  & 9  & \textbf{1}  & \textbf{4}  & \textbf{2}  & 6 \\\hline
\end{tabular}}{}
\end{table*}
In Table \ref{tab:attribute_ranking}, we record the ranking of the used attributes in our experiments. For more generalization, RFE was performed on each of the six datasets using a combination of each of the top-five distance measures and each of the top-five values of $k$. The total number of RFE experiments is 150. For each attribute, we count the total number of times it appeared in the optimal subset of attributes. A score of $\frac{total\textit{ }count}{number\textit{ }of\textit{ }experiments}$ is assigned to each attribute according to its total count. It is clear that the best subset of attributes is dataset dependent. The five most informative attributes are respectively: A15 (energy), A17 (link impurity), A12 (number of distinct eigenvalues), A16 (neighborhood impurity), and A13 (spectral radius). All spectral attributes showed to be very informative. Indeed, three of them (A15, A12, and A13) ranked in the top-five, and A14 (second largest eigenvalue) ranked in the top-ten (9$^{th}$) with a score of 0.52 meaning that for more than half of all the experiments, all spectral attributes were selected in the optimal subset of attributes. Unsurprisingly, A11 (percentage of end points) ranked last with a very low score. This is because proteins are dense molecules and thus very few nodes of their respective graphs will be end points (extremity amino acids in the primary structure with no spatial links). Label attributes also showed to be very informative. Indeed, A17, A16, and A18 (label entropy) ranked respectively 2$^{nd}$, 4$^{th}$, and 6$^{th}$ with scores of more than 0.61. This is due to the importance of the distribution of the types of amino acids and their interactions. Both have to follow a certain harmony in the structure in order to exert a particular function. A9 (closeness centrality), A5 (average clustering coefficient) and A8 (effective radius) ranked in the top-ten with scores of more than 0.5 (A8 scored 0.49 $\simeq$ 0.5). However, all A1 (number of nodes), A2 (number of edges), A3 (average degree), A4 (density), A6 (average effective eccentricity), A7 (effective diameter), and A10 (percentage of central nodes) scored less than 0.5. This is because each one of them is represented by one of the top-ten attributes and thus presents a redundant information. A6 and A9 are both expressed based on all shortest paths of the graph. Both A7 and A8 are expressed based on A6. A10 is expressed based on A8 and thus on A6 too. A1, A2, A3, and A4 are all highly correlated to A5.

\subsubsection{Comparison with Other Classification Techniques}
We compare our approach with multiple state-of-the-art approaches for protein function prediction namely: sequence alignment-based classification (using Blast \citep{Altschul_1990}), structural alignment-based classification (using Combinatorial Extension (CE) \citep{Bourne_1998}, Sheba \citep{Lee_2000}, and FatCat \citep{Ye_2003}), and substructure(subgraph)-based classification (using GAIA \citep{Jin_2010}, LPGBCMP \citep{Fei_2010}, and D\&D \citep{Zhu_2012}). For sequence and structural alignment-based classification, we align each protein against all the rest of the dataset. We assign to the query protein the function of the reference protein with the best hit score. For the substructure-based approaches, all the selected approaches are mainly for mining discriminative subgraphs. LPGBCMP is used with $max_{var}$ = 1 and $d$ = 0.25 for, respectively, feature consistency map building and overlapping. In \citep{Fei_2010}, LPGBCMP outperformed several other approaches from the literature including LEAP \citep{Yan_2008}, gPLS \citep{Saigo_2008}, and COM \citep{Jin_2009} on the classification of the same six benchmark datasets. GAIA showed in \citep{Jin_2010} that it outperformed other state-of-the-art approaches namely COM and graphSig \citep{Ranu_2009}. D\&D have showed in \citep{Zhu_2012} that it also outperformed COM and graphSig, and that it is highly competitive to GAIA. 
For all these approaches, the discovered substructures are considered as features for describing each example of the original data. The constructed description matrix is used for training in the classification. For our approach, we show the classification accuracy results of \textsc{ProtNN} with RFE using std-Euclidean distance. We also show the best results of \textsc{ProtNN} (denoted \textsc{ProtNN}*) with RFE using each of the top-five distance measures. We use $k$ = 1 both for \textsc{ProtNN} and \textsc{ProtNN}*. Table \ref{tab:comparison} shows the obtained results. 

The alignment-based approaches FatCat and Sheba outperformed CE, Blast, and all the subgraph-based approaches. Indeed, FatCat scored best with three of the first four datasets and Sheba scored best with the two last datasets. 
Except CE, all the other approaches scored on average better than Blast. This shows that the spatial information constitutes an important asset for functional classification by emphasizing structural properties that the primary sequence alone do not provide. 
For the subgraph-based approaches, D\&D scored better than LPGBCMP and GAIA on all cases except with DS1 where GAIA scored best. 
On average, \textsc{ProtNN*} ranked first with the smallest distance between its results and the best obtained accuracies with each dataset. This is because \textsc{ProtNN} considers both structural information, and hidden topological properties that are omitted by the other approaches.

\begin{table*}[]
\processtable{Accuracy comparison of \textsc{ProtNN} with other classification techniques.\label{tab:comparison}}
{\begin{tabular}{llllllllll}
\hline\noalign{\smallskip}
\multicolumn{1}{c}{\multirow{2}{*}{\textbf{Dataset}}} & \multicolumn{9}{c}{\textbf{Classification approach}} \\ \cline{2-10}\noalign{\smallskip}
\multicolumn{1}{c}{} & \textbf{Blast} & \textbf{Sheba} & \textbf{FatCat} & \textbf{CE}  & \textbf{LPGBCMP} & \textbf{D\&D}  & \textbf{GAIA}  & \textbf{\textsc{ProtNN}} & \textbf{\textsc{ProtNN*}}\\ 
\hline\noalign{\smallskip}
DS1     & 0.88 & 0.81 & \textbf{1}    & 0.45 & 0.88   & 0.93    & \textbf{1}   & 0.97    & 0.97\\
DS2     & 0.82 & 0.86 & \textbf{0.89}  & 0.49 & 0.73   & 0.76    & 0.66    & 0.8     & \textbf{0.89}\\
DS3     & 0.9 & 0.95 & 0.84  & 0.59 & 0.90   & 0.96    & 0.89    & 0.96    & \textbf{0.97}\\
DS4     & 0.76 & 0.92 & \textbf{1}    & 0.46 & 0.9   & 0.93  & 0.89    & 0.97    & 0.97\\
DS5     & 0.86 & \textbf{0.99} & 0.94  & 0.76 & 0.87   & 0.89  & 0.72    & 0.9       & 0.94\\
DS6     & 0.78 & \textbf{1}   & 0.94   & 0.81 & 0.91   & 0.95    & 0.87    & 0.96    & 0.96\\ \hline\noalign{\smallskip}
\textbf{Avg. accuracy}$^1$ & 0.83$\pm$0.05 & 0.92$\pm$0.07 & 0.94$\pm$0.06 & 0.59$\pm$0.15 & 0.86$\pm$0.06 & 0.9$\pm$0.07 & 0.84$\pm$0.12 & 0.93$\pm$0.06 & \textbf{0.95$\pm$0.03}\\ \hline\noalign{\smallskip}
\textbf{Avg. distances}$^2$ & 0.14$\pm$0.07 & 0.05$\pm$\small{0.07} & 0.04$\pm$0.05 & 0.38$\pm$0.15 & 0.11$\pm$0.03 & 0.7$\pm$0.04 & 0.14$\pm$0.09 & 0.05$\pm$0.03 & \textbf{0.02$\pm$0.01}\\ \hline\noalign{\smallskip}
\textbf{Rank}    &   8   &   4   &   2   &   9   &   6   &   5   &   7   &   3   &  \textbf{1}\\ \hline\noalign{\smallskip}
\end{tabular}}{$^1$Average classification accuracy of each classification approach over the six datasets.\\
$^2$Average of the distances between the accuracy of each approach and the best obtained accuracy with each dataset.}
\end{table*}

\subsection{Scalability and Runtime Analysis}
\begin{figure}[!htpb]
\centerline{\includegraphics[width=0.5\textwidth]{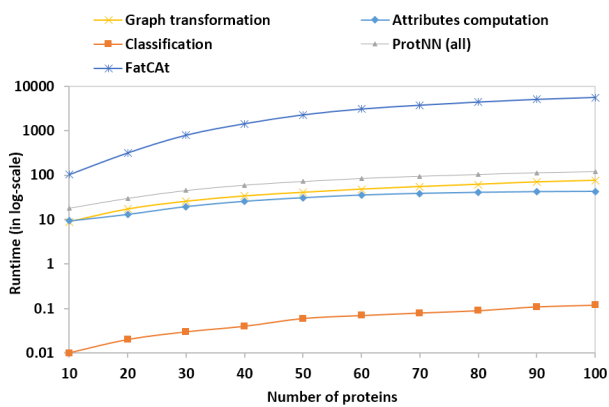}}
\caption{Runtime comparison in log-scale of \textsc{ProtNN} and FatCat. The running time of \textsc{ProtNN} is separated for the main steps. }\label{fig:runtime}
\end{figure}

Besides being accurate, an efficient approach for functional classification of protein 3D-structures has to be very fast in order to provide practical usage that meets the increasing load of data in real-world applications. In this section, we study the runtime of \textsc{ProtNN} and FatCat, the most competitive approach in our previous comparative experiments. We analyze the variation of runtime for both approaches with higher numbers of proteins ranging from 10 to 100 3D-structures with a step-size of 10. In Figure \ref{fig:runtime}, we report the runtime results in log$_{10}$-scale. A huge gap is clearly observed between the runtime of \textsc{ProtNN} and that of FatCat. The gap gets larger with higher numbers of proteins. Indeed, FatCat took over 5570 seconds with the 100 proteins while \textsc{ProtNN} (all) did not exceed 118 seconds for the same set which means that our approach is 47x faster than FatCat on that experiment. The average runtime of graph transformation of \textsc{ProtNN} was 0.8 second and that of the computation of attributes was 0.6 second for each protein. The total runtime of similarity search and function prediction of \textsc{ProtNN} was only 0.1 on the set of 100 proteins. Note that in real-world applications, the preprocessing (graph transformation and attribute computation) of the reference database is performed only once and the latter can be updated with no need to recompute the existing values. This ensures computational efficiency and easy extension of our approach. 

\subsubsection{Scalability to a PDB-wide classification}
We further evaluate the scalability of \textsc{ProtNN} in the classification of the entire Protein Data Bank (described in \ref{subsec:PDB}). We also show the runtime for FatCat and CE (the structural alignment approaches used in the PDB website\footnote{http://www.rcsb.org/pdb/}). We recall that the experiments are on a single process mode with no parallelization for all the approaches. Note that in the PDB website, the structural alignment is whether pre-computed for structures of the database, or only performed on a sub-sample of the PDB for customized or local files. Table \ref{tab:runtime_on_PDB} shows the obtained results. It is clear that the computation of attributes is the most expensive part of our approach as some attributes are very complex. However, building the graph models and the computation of attributes represent the preprocessing step and are only performed once for the reference database. The classification step took almost three hours with an average runtime of 0.1 second for the classification of each protein against the entire PDB. All and all \textsc{ProtNN} runtime was less than a week with an average runtime of 5.9 seconds for the preprocessing and classification of each protein 3D-structure against the entire PDB. On the other hand, both FatCat and CE did not finish running within two weeks. We computed the average runtime for each approach on the classification of a sample of 100 proteins against all the PDB. On average FatCat and CE took respectively more than 42 and 32 hours per protein making our approach faster than both approaches with thousands orders of magnitude on the classification of a 3D-structure against the entire PDB.

\begin{table}[!t]
\processtable{Runtime results of \textsc{ProtNN}, FatCat and CE on the entire Protein Data Bank.\label{tab:runtime_on_PDB}}
{\begin{tabular}{lll}
\hline\noalign{\smallskip}
\textbf{Task} & \textbf{Total runtime$^1$} & \textbf{Runtime$^1$/protein}\\ 
\hline\noalign{\smallskip}
Building graph models & 23h:9m:57s & 0.9s\\
Computation of attributes & 5d:8h:12m:29s & 4.9s\\
Classification & 2h:55m:15s & 0.1s\\
\textbf{\textsc{ProtNN} (all)} & 6d:10h:17m:41s & 5.9s\\\hline
\textbf{\textsc{FatCat}} & Forever$^2$ & 1d:18h:31m:35s$^3$\\\hline
\textbf{\textsc{CE}} & Forever$^2$ & 1d:8h:37m:34s$^3$\\\hline
\end{tabular}}{$^1$The runtime is expressed in terms of days:hours:minutes:seconds\\
$^2$The program did not finish running within two weeks\\
$^3$The average runtime of randomly selected 100 proteins}
\end{table}

\section{Conclusion}
In this paper, we proposed \textsc{ProtNN}, a new fast and accurate approach for protein function prediction. We defined a graph transformation and embedding model that incorporates explicit as well as hidden structural and topological properties of the 3D-structure of proteins. We successfully implemented the proposed model and we experimentally demonstrated that it allows to detect similarity and to predict the function of protein 3D-structures efficiently. Empirical results of our experiments showed that considering structural information constitutes a major asset for accurately identifying protein functions. They also showed that the alignment-based classification as well as subgraph-based classification present very competitive approaches. Yet, as the number of pairwise comparisons between proteins grows tremendously with the size of dataset, enormous computational costs would be the results of more detailed models. Here we highlight that \textsc{ProtNN} could accurately classify multiple benchmark datasets from the literature with very low computational costs. With all large-scale studies, it is an asset that  \textsc{ProtNN} scale up to a PDB-wide dataset in a single-process mode with no parallelization, where it outperformed state-of-the-art approaches with thousands order of magnitude in runtime on classifying a 3D-structure against the entire PDB.
In future works, we aim to integrate more proven protein functional attributes in our model to further enhance the accuracy of the prediction system.  

\section*{Acknowledgements}
This study is funded by the Natural Science and Engineering Research Council through a discovery grant to ABD.
%
%
\\
\paragraph{Conflict of Interest:} none declared.
\\
\bibliographystyle{natbib}
%
%
\bibliography{Thesis}
\end{document}